\pdfoutput=1

\documentclass[11pt]{article}

\usepackage[final]{acl}

\usepackage{times}
\usepackage{latexsym}
\usepackage{booktabs} 
\usepackage{tabularx}
\usepackage{ragged2e}
\usepackage{stfloats}
\usepackage[dvipsnames]{xcolor}
\usepackage{hyperref}
\usepackage{xurl}
\usepackage{multirow}
\usepackage{float}
\usepackage{caption}
\usepackage[T1]{fontenc}

\usepackage[utf8]{inputenc}

\usepackage{microtype}

\usepackage{inconsolata}

\usepackage{graphicx}

\title{Where Are Language-Specific Features Represented in Multilingual LLMs? Everywhere}

\author{
  Mikaela Martins, Jonathan Dunn \\
  Department of Linguistics \\
  University of Illinois Urbana-Champaign \\
  \texttt{mlm25@illinois.edu} \quad
  \texttt{jedunn@illinois.edu}
}

\begin{document}
\maketitle
\begin{abstract}
Previous work has argued that multilingual LLMs encode language-specific features most strongly in their initial and final layers, implying that intermediate layers support language-agnostic representations. This paper reconsiders this finding by performing four different experiments: three that undertake a fine-grained analysis of representations of Portuguese and Spanish, two closely related languages, and one large-scale language identification experiment that covers over 200 languages. The results of these experiments challenge the strict separation between language-specific and language-agnostic layers in multilingual LLMs, using Llama 3.2 1B as a test case. By analyzing and manipulating contextual sentence embeddings, we demonstrate that language identity remains strongly recoverable throughout the model rather than being restricted to initial and final layers. Moreover, we find that the model tends to classify ambiguous strings as closer to Spanish than to Portuguese, revealing a skew in model behavior that differs from human judgments. 
\end{abstract}

\section{Introduction}
\label{sec:intro}

This paper experiments with how LLMs represent language-specific information in their embeddings by focusing on strings that display \textit{language membership ambiguity} between closely related languages. We define language membership ambiguity as a condition in which the surface form of a string is compatible with more than one language. We contrast \textit{ambiguous} strings, which are compatible with both Portuguese and Spanish, with \textit{non-ambiguous} strings that contain sufficient language-specific cues to be identified as belonging to a specific language. We use this contrast as a means of identifying to what degree embeddings from different layers encode language-specific information. We then extend this investigation to over 200 different languages with a classification experiment.

While there has been much work on multilingual representations, this problem remains only partially understood \cite{chang2022, tang-etal-2024, lindsey2025biology}. Contextual embeddings have been shown to combine both language-specific and language-agnostic information \cite{chen2023, dumas2024, ferrando2024, fierro2025, tamo2026}. The idea is that, in some layers, embeddings contain more language-specific content \cite{liang2021, tang-etal-2024, salim2026}: therefore, these embeddings should differ between ambiguous and non-ambiguous strings. But other layers are expected to contain language-agnostic content \cite{wendler2024, dumas2024, wu2025semantichubhypothesislanguage, dumas2025}, for example, with reasoning processes taking place in English \cite{zhao2024, schut2025, lindsey2025biology}. Thus, ambiguous and non-ambiguous strings should not differ in these layers. Our first three experiments test this contrast. By extending the investigation beyond Portuguese and Spanish in the fourth experiment, we assess whether these observations generalize to sets of languages that are not closely related. Our focus is on the aggregate representation of language-specific information across layers rather than identifying circuits or neurons responsible for these features. 

Understanding how multilingual LLMs store and organize linguistic information is important because failures to distinguish between languages have observable consequences in multilingual systems. \citet{chen-etal-2023-target} describe the \textit{off-target language issue} in machine translation, in which a model generates a translation in the wrong language. The authors associate this behavior with a failure to encode sufficiently discriminative target-language signals and with greater lexical similarity between languages. Similarly, \citet{marchisio-etal-2024-understanding} use the term \textit{language confusion} for cases in which LLMs fail to consistently generate text in the contextually appropriate language. Language membership ambiguity is not itself such an error; rather, it provides a controlled setting in which surface form alone does not uniquely determine language identity, allowing us to examine how multilingual models internally represent language membership. Moreover, understanding these properties of multilingual representations is also important for understanding how multilingual LLMs handle interlingual phenomena such as code-switching, code-mixing, and dialectal variation within languages, where language identity itself becomes fuzzy.  

This paper has two main goals: (i) to measure how strongly language-specific information is encoded across layers of multilingual LLMs. And (ii) to map how these models represent related languages that display language membership ambiguity in their embeddings. We use Portuguese and Spanish as a test case because their shared roots make them very similar in structure and vocabulary: the number of cognates between Spanish and Portuguese reaches 85\% \cite{ulsh}. Other language pairs that display language membership ambiguity include Indonesian and Malay or Croatian and Serbian \cite{goutte-etal-2016}. Similar languages are thus most likely to contain strings which are ambiguous, being interpretable in multiple languages.

This study is divided into four experiments: the first outlines the multilingual representation space of Llama 3.2 1B using parallel sentences. The second examines how the model represents ambiguous strings, where language identity is not strictly constrained. The third investigates whether language-specific priming can shift these representations and reduce ambiguity. Finally, the fourth trains a language identification classifier on sentence-level embeddings from each layer for over 200 languages. Examples of ambiguous and non-ambiguous strings are found in Table \ref{table1} and in Appendix D.  

When interpreted together, these experiments provide a fine-grained test case involving two similar languages and a broad analysis of multilingual representations across many languages and layers for the sake of determining what regions of LLMs store language-specific representations. Our results show that language-specific information remains strongly recoverable throughout the model, regardless of layer. We also find that the model tends to classify ambiguous strings as closer to Spanish, showing a default preference. Adding context to ambiguous structures helps assess this skew.

\section{Reasoning and Language Use in LLMs}
\label{sec:litrev}


Language-specific features are predominantly reported in the first and last (or top and bottom) layers of multilingual LLMs \cite{tang-etal-2024, liang2021, lindsey2025biology, wendler2024, salim2026}. \citet{tang-etal-2024} found \textit{language-specific neurons} within LLMs: neurons whose activation distribution differs systematically across languages. They demonstrate that similar languages (i.e., Japanese and Chinese) or languages of the same family share many language-specific neurons: around 25\% of neurons activated to solve a task related to Japanese also activate to solve tasks related to Chinese and, as a result, when deactivating the neurons specific to one language, the performance of the other language is impacted \cite{tang-etal-2024, zhao2024}. Furthermore, they found that all languages appear to have an equal number of language-specific neurons, except for English, which has significantly fewer. They hypothesize that this happens because English is the most well represented language in the training data of most LLMs, becoming the default language.

Language-agnostic regions have been located in intermediate layers \cite{chang2022, chen2023, ferrando2024, lindsey2025biology}, but they are mostly aligned towards English. This led some authors to claim that English is the language used for reasoning in middle layers \cite{schut2025, zhao2024, tamo2026}. \citet{zhao2024} notes that, initially, LLMs convert multilingual inputs into English for the sake of reasoning. Once the intermediate layers have solved the task, later layers produce an output that matches the original language. They also found that, in Chinese Llama, Chinese is the primary language used for reasoning across all languages. Thus, they propose that the dominant language in the training data dictates the model's default reasoning language \cite{zhao2024, zhong2024}. 

However, it is also argued that intermediate layers are organized around concepts \cite{wendler2024, dumas2024, brinkmann-etal-2025-large, fierro2025, wu2025semantichubhypothesislanguage, dumas2025}; that is, LLMs operate in an abstract conceptual space. In this view, the embeddings’ proximity to English follows from an English bias in the concept space itself, rather than from the model first translating inputs to English for the sake of reasoning. \citet{wu2025semantichubhypothesislanguage} found that LLMs have a shared representational space that stores data types beyond language itself, even without explicit alignment transformation. This space is scaffolded by the predominant language in the training data (which is usually but not always English). 



Given this distinction between language-specific and language-agnostic regions in multilingual LLMs, we expect closely related languages to share more representational space, particularly when language membership is ambiguous. Accordingly, we expect distinct languages to form clearer language-specific clusters in the initial and final layers, where previous work has located the majority of language-specific information, but to show greater overlap in intermediate layers, where representations are expected to be more language-agnostic and potentially organized around concepts. Ambiguous strings should therefore fall between the Portuguese and Spanish clusters in more language-specific layers, while their representations should be less strongly associated with either language in intermediate layers. Finally, we expect language classification performance to be higher in the initial and final layers than in intermediate ones.

\section{Experiment 1: Representing Parallel Sentences} 
\label{sec:exp1}

The first experiment analyzes contextual embeddings for parallel sentences to determine if they are organized by language or by content. If certain layers of the model are language-agnostic, then the content-based organization should be strongest. This experiment uses three Romance languages: Spanish vs Portuguese with French as a control language, in the sense that French should be far separated from the other two. Equivalent sentences are drawn from the Europarl corpus and then the experiment is replicated with parallel sentences from TED Talks to ensure that the findings are robust across registers (they are).

\textbf{Model}. For this and all following experiments, we use Llama 3.2 1B\footnote{https://huggingface.co/meta-llama/Llama-3.2-1B} \cite{llama32_1b}, an autoregressive decoder-only transformer. This model has 16 layers with 2048 dimensions each. We report results for layers 1, 8, and 16 in the main text of the paper; the appendices contain complete results for all layers. This model was trained for many languages, but only eight are officially supported: English, German, French, Italian, Portuguese, Hindi, Spanish, and Thai. To extract sentence embeddings from the model, we use mean-pooling with an attention mask and average the embeddings for each token. The attention mask excludes the padding tokens from the average, ensuring that only valid linguistic tokens contribute to the final representation. We then standardize the sentence embeddings independently for each layer. 

\begin{figure*}[!t]
    \centering
    \includegraphics[width=0.48\textwidth]{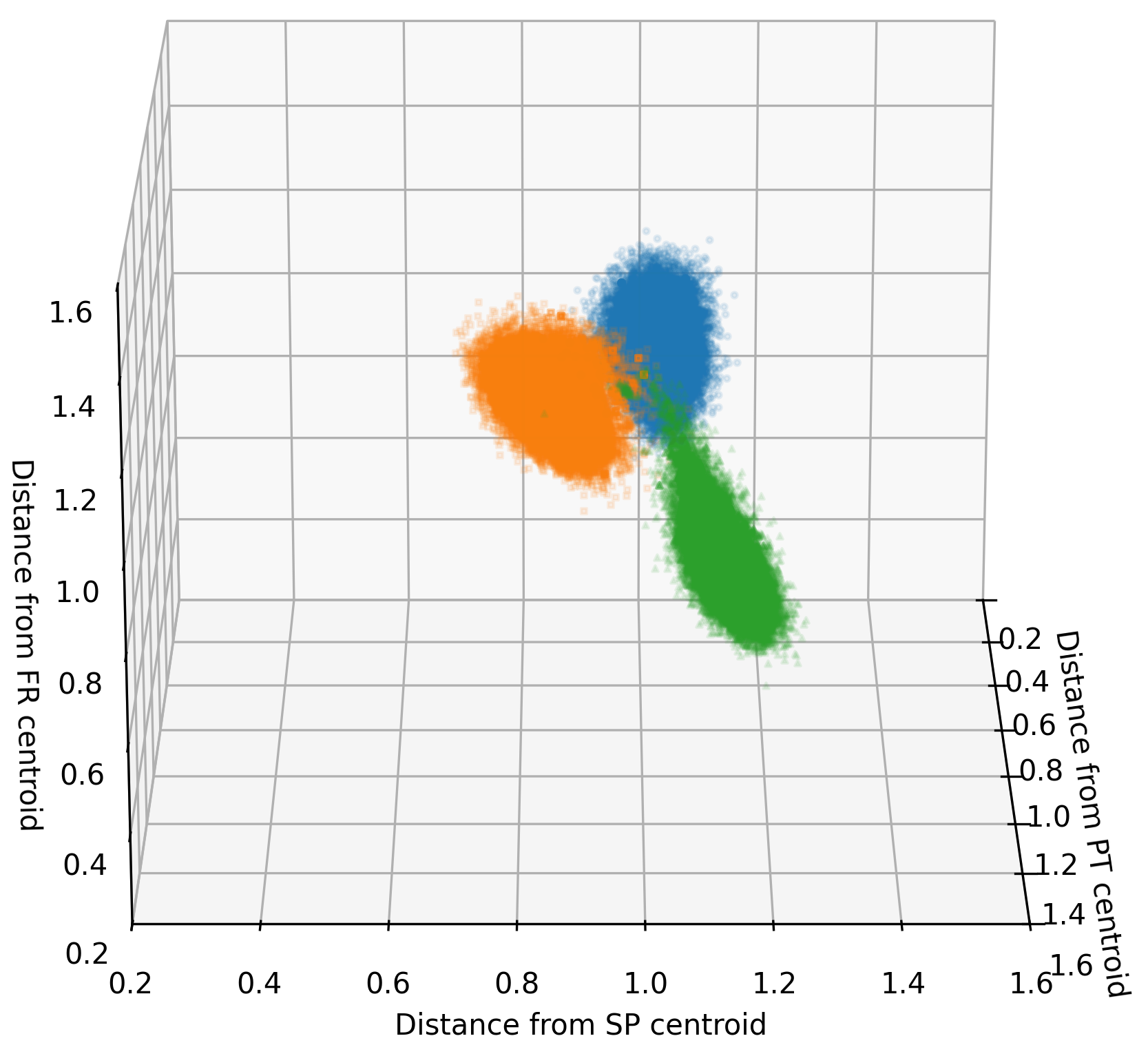}
    \hfill
    \includegraphics[width=0.48\textwidth]{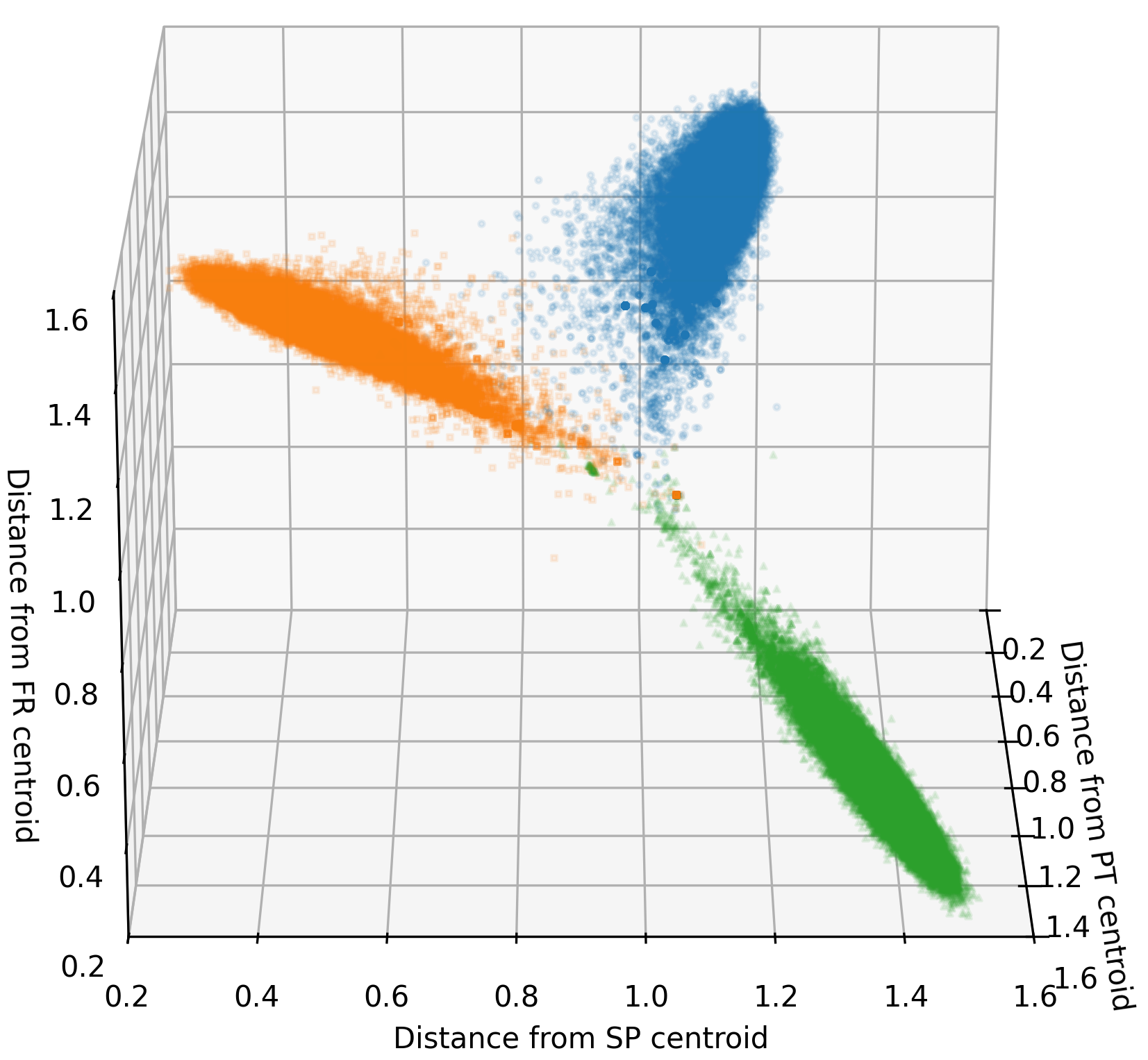}
    \caption{Language-specific clusters for the Europarl Corpus in layer 8 (left) and layer 16 (right) of Llama 3.2 1B. Each color represents a different language: Portuguese (blue), Spanish (orange), and French (green). The x, y, and z axes display the raw cosine distances from the Spanish, French, and Portuguese centroids, respectively.}
    \label{fig1}
\end{figure*}

\textbf{Data}. This experiment uses the Spanish, Portuguese, and French sections of the Europarl Corpus\footnote{https://opus.nlpl.eu/Europarl/corpus/version/Europarl}, taken from the OPUS Collection\footnote{https://opus.nlpl.eu/} \cite{opus}. In total, the experiment uses 400,000 sentences, each found in all three languages. The goal of the parallel corpus is to position sentences with the same content in different languages in order to measure how much of the information encoded in the embeddings is language-specific. The only difference between two parallel sentences should be the language-specific information. Because parliamentary proceedings are a very conventionalized register, we then replicate the same experiment on a second corpus: the TED2020 Corpus\footnote{https://opus.nlpl.eu/TED2020/corpus/version/TED2020}, a collection of text that contains nearly 4,000 TED and TEDx transcripts. We again use Spanish and Portuguese as the languages of interest, with French as a control language.

The question here is the degree to which there is language-specific information in the sentence embeddings extracted from each layer of the model. For each language, we calculate its centroid by taking the mean of all sentence embeddings belonging to that language. This centroid thus captures the approximate location of Spanish-ness or Portuguese-ness in the sentence embeddings. For each parallel sentence, we then calculate its cosine distance to the three language centroids (Spanish, Portuguese, French). If a particular layer is language-agnostic, we would expect parallel versions of the same sentence to be equally distant from each language centroid. If, on the other hand, a particular layer contains language-specific information, we would expect that, for instance, the Spanish version would be close to the Spanish centroid.

The distances between each parallel sentence and the language-specific centroids are plotted in Figure \ref{fig1} with specific colors for each language. Each point is a sentence, and the three axes represent the distance from each centroid. A language-agnostic representation would create a plot in which all the colors were overlapping. Instead, we see that there are three clear language-specific clusters. The Spanish (orange) and Portuguese (blue) clusters are much closer to one another than to the French control cluster (green). There is some overlap between clusters in Layer 8 (the left plot) but much less overlap in Layer 16 (the right plot). This perhaps corresponds with earlier work showing that intermediate layers contain less language-specific information than other layers \cite{chang2022, zhao2024, tang-etal-2024}. However, Layer 8 still shows strong language separation, even if the clusters are not as distinct as in Layer 16. This is initial evidence that language-specific information is relevant and used by the model, even in layers where previous work has argued there is less language-specific information.

\begin{table*}[!b]
\centering
\caption{Distinctiveness of language-specific clusters in the Europarl Corpus. \textit{Portuguese}, \textit{Spanish}, and \textit{French} at the top refer to the language-specific clusters; thus, the first column represents how close Portuguese sentences are to other Portuguese sentences relative to their parallel equivalents. The \textit{L} column refers to the layer. The values under \textit{Mean distances} indicate the actual distance to each centroid, standardized relative to the global mean and standard deviation in that layer. Thus, a value of -1.2 would indicate that Portuguese sentences in layer one are more than one standard deviation closer to other Portuguese sentences than to their parallel counterparts. Both measures indicate a consistent level of language-specific information within each layer.}
\label{table2}
\scriptsize
\begin{tabular}{l cccccc||cccccc||ccccccc}
\toprule
 & \multicolumn{6}{c}{\textbf{Portuguese}} 
 & \multicolumn{6}{c}{\textbf{Spanish}} 
 & \multicolumn{6}{c}{\textbf{French}} \\
\cmidrule(lr){2-7} \cmidrule(lr){8-13} \cmidrule(lr){14-19}
\textbf{L} 
& \multicolumn{3}{c}{Nearest centroid (\%)} 
& \multicolumn{3}{c}{Mean distances}
& \multicolumn{3}{c}{Nearest centroid (\%)} 
& \multicolumn{3}{c}{Mean distances}
& \multicolumn{3}{c}{Nearest centroid (\%)} 
& \multicolumn{3}{c}{Mean distances} \\
\cmidrule(lr){2-4} \cmidrule(lr){5-7}
\cmidrule(lr){8-10} \cmidrule(lr){11-13}
\cmidrule(lr){14-16} \cmidrule(lr){17-19}
& PT & SP & FR & PT & SP & FR
& PT & SP & FR & PT & SP & FR
& PT & SP & FR & PT & SP & FR \\
\midrule

\textbf{1} 
& 99 & 0.8 & 0.5 & -1.2 & 0.2 & 0.9
& 0.0  & 99 & 0.0 & 0.2 & -1.1 & 0.9
& 0.0  & 0.0 & 99 & 0.8 & 0.7 & -1.4 \\

\textbf{8} 
& 99 & 1.5 & 0.8 & -1.1 & 0.1 & 1.1
& 0.1  & 98 & 0.0 & 0.1 & -1.0 & 0.9
& 0.2  & 0.1 & 99 & 0.8 & 0.5 & -1.5 \\

\textbf{16} 
& 99 & 0.0 & 0.0 & -1.2 & 0.3 & 0.9
& 0.3  & 99 & 0.2 & 0.3 & -1.2 & 0.9
& 0.5  & 0.1 & 99 & 0.8 & 0.7 & -1.4 \\

\bottomrule
\end{tabular}
\end{table*}

Table \ref{table2} contains more detailed information about the distinctiveness of language-specific clusters in the embedding space. For each Portuguese sentence, if it is closer to other random Portuguese sentences than to its counterpart in the other languages, then it counts as belonging to the Portuguese cluster. As shown in the table, most sentences are organized by language in this way (\textit{Nearest centroid} in the table). Similarly, the average distance is also given: French, as expected, is even more distinctive as a cluster because its sentences are much less similar than Spanish and Portuguese (\textit{Mean distances} in the table). Hence, in these layers, all languages demonstrate well-defined clusters, particularly because all sentences sit more than one standard deviation inside each cluster on average. A more detailed breakdown across all layers is available in Appendix A; the results shown here are representative of the overall trend. While Appendix A provides a layer-by-layer breakdown for the Europarl Corpus, Appendices B and C repeat the analysis for the TED Corpus. These results also mirror the findings presented here. 

The question in this first experiment is whether sentence embeddings for parallel corpora are organized by language (indicating the presence of more language-specific information) or by content (indicating the presence of more language-agnostic information). While there are small variations, with less language-specific information in intermediate layers, there is a consistent trend that all layers are dominated by linguistic rather than content-based information. Our interim conclusion is that sentence embeddings for all layers contain significant language-specific information because language identity remains a major organizing factor in sentence-level representations across all layers. Even when sentences express equivalent content, Portuguese, Spanish, and French form distinct language-specific clusters. 

\begin{figure}[!b]
    \centering
    \includegraphics[width=0.5\textwidth]{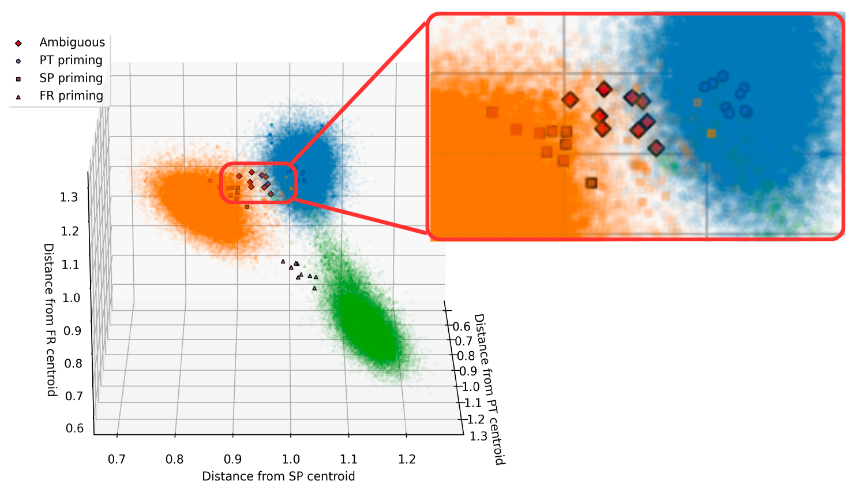}
    \caption{Distribution of ambiguous and non-ambiguous strings in layer 8. The blue, orange, and green clusters represent Portuguese, Spanish, and French, respectively. The x, y, and z axes display the raw cosine distances from the Spanish, French, and Portuguese centroids in the Europarl corpus, respectively. The outlined red diamonds represent the ambiguous sentences, located between the Portuguese and Spanish clusters, enlarged in the red box.}
    \label{fig2}
\end{figure}

\begin{table*}[!t]
\centering
\setlength{\tabcolsep}{4pt}          
\renewcommand{\arraystretch}{1.15}   
\footnotesize                        
\begin{tabularx}{\textwidth}{
  >{\RaggedRight\arraybackslash}X
  >{\RaggedRight\arraybackslash}X
  >{\RaggedRight\arraybackslash}X
  >{\RaggedRight\arraybackslash}X}
\toprule
\textbf{Ambiguous} &
\textbf{Portuguese} &
\textbf{Spanish} &
\textbf{French} \\
\midrule
Este documento será publicado esta semana. [\textit{This document will be published this week.}] &
\textcolor{Red}{Comuniquei-o ontem a respeito disso.} Este documento será publicado esta semana. [\textit{\textcolor{Red}{I informed him about this yesterday.} This document will be published this week.}] &
\textcolor{Red}{No hay escapatoria.} Este documento será publicado esta semana. [\textit{\textcolor{Red}{There's no way out.} This document will be published this week.}] &
\textcolor{Red}{Il n’y a aucune issue.} Este documento será publicado esta semana. [\textit{\textcolor{Red}{There is no way out.} This document will be published this week.}] \\
\bottomrule
\end{tabularx}
\caption{Example from the Ambiguous String Corpus. The first column contains an ambiguous string that shows language membership ambiguity between Portuguese and Spanish. The second, third, and fourth columns exemplify the structure of non-ambiguous strings with priming in Portuguese, Spanish, and French, respectively. French was chosen to represent the neutral condition given that it does not display language membership ambiguity with the other two languages. In the non-ambiguous condition, the priming structure is highlighted in red; the rest of the string is identical to the ambiguous sentence. English translations are provided between square brackets. More examples are found in Appendix D.}
\label{table1}
\end{table*}

\section{Experiment 2: Representing Ambiguous Strings}

The second experiment repeats this methodology for strings that are ambiguous between Spanish and Portuguese: these are strings which, for a model without more context, could belong to either language. We use these ambiguous structures to determine if the model has a skew that pushes it to view potentially Portuguese strings as Spanish strings, indicating a preference for Spanish, or if it has a skew that pushes it to view potentially Spanish strings as Portuguese strings, indicating a preference for Portuguese.

\textbf{Data}. We construct a corpus of ambiguous strings (Table \ref{table1}). This corpus contains seven distinct types of ambiguous constructions to ensure variability, with 10 sentences per category. Thus, the ambiguous corpus has 70 strings for which a model without context could not attribute them to either Spanish or Portuguese. All strings were collected from news sources such as \textit{El País}, \textit{CNN en Español}, \textit{Portal G1}, and \textit{O Estadão} and were reviewed by (i) native monolingual Spanish speakers from Spain and Mexico, and (ii) native monolingual Portuguese speakers from Brazil, who unanimously confirmed their language membership. Each ambiguous string is paired with a non-ambiguous string for each language to support a contrast between ambiguous and non-ambiguous strings. French continues to act as a control language. The strings were adapted to remove named entities and include proper nouns used in both Portuguese and Spanish-speaking countries. 

We use the same model and the same method for extracting sentence embeddings as in Experiment 1. We continue to use the sentences from Europarl as an independent measure of the centroid for each language: where we would expect sentences to fall if they belonged entirely to one or the other language. The goal is to observe how the model views these strings. Ambiguous strings closer to the Portuguese centroid mean that the model believes these strings to be more similar to Portuguese. Strings located between the two clusters and further away from both centroids indicate that the model views them as ambiguous.

Figure \ref{fig2} and Table \ref{table3} position these new strings against norms provided by Europarl. The figure shows Layer 8; ambiguous strings are the red diamonds located between the two clusters, while the non-ambiguous strings are positioned within each language-specific cluster. In most layers, the ambiguous strings sit between the Portuguese and Spanish clusters, as seen in Figure \ref{fig2}. Table \ref{table3}, however, tells a more nuanced story: ambiguous strings are closer to the Portuguese centroid in the initial layers, but start moving closer to the Spanish centroid in layer 4 (not shown here; c.f., Appendix E). This suggests that the model tends to see ambiguous strings as closer to Spanish than Portuguese, which implies that these strings are not fully ambiguous to the model the way they are to humans. 

This experiment used ambiguous strings as a test case to determine whether human judgments of language membership ambiguity are aligned with machine judgments on the same phenomenon: if so, we would expect that strings which could belong to either language end up falling between the language-specific clusters. This is exactly what we find in earlier layers, but the results suggest that later layers increasingly resolve ambiguous forms in a Spanish-oriented direction. 

\begin{table*}[!b]
\centering
\caption{Ambiguous String Corpus: bold \textit{Portuguese}, \textit{Spanish}, and \textit{French} at the top of each section refer to the language clusters. The \textit{L} column refers to layer number. The upper section contains the percentages of strings in each language that are closer to the Portuguese, Spanish, and French centroids, respectively. The lower section contains their average distance to each centroid. Distances are standardized.}
\label{table3}
\scriptsize
\begin{tabularx}{\textwidth}{l *{12}{>{\centering\arraybackslash}X}}
\toprule
 & \multicolumn{4}{c}{\textbf{Portuguese}} 
 & \multicolumn{4}{c}{\textbf{Spanish}} 
 & \multicolumn{4}{c}{\textbf{French}} \\
\cmidrule(lr){2-5} \cmidrule(lr){6-9} \cmidrule(lr){10-13}
\textbf{L} 
& PT & SP & FR & AB
& PT & SP & FR & AB
& PT & SP & FR & AB \\
\midrule
\textbf{1}  & 100.0 & 8.45 & 5.63 & 84.51  & 0.00 & 91.55 & 78.87 & 15.49  & 0.00 & 0.00 & 15.50 & 0.00 \\
\textbf{8}  & 100.0 & 1.41 & 0.00 & 77.46  & 0.00 & 98.59 & 0.00 & 22.54  & 0.00 & 0.00 & 100.0 & 0.00 \\
\textbf{16} & 2.82 & 0.00 & 0.00 & 0.00  & 0.00 & 98.59 & 0.00 & 32.39  & 97.18 & 1.41 & 100.0 & 67.61 \\
\bottomrule
\end{tabularx}

\vspace{0.2em}

\begin{tabularx}{\textwidth}{l *{12}{>{\centering\arraybackslash}X}}
 & \multicolumn{4}{c}{\textbf{Portuguese}} 
 & \multicolumn{4}{c}{\textbf{Spanish}} 
 & \multicolumn{4}{c}{\textbf{French}} \\
\cmidrule(lr){2-5} \cmidrule(lr){6-9} \cmidrule(lr){10-13}
\textbf{L} 
& PT & SP & FR & AB
& PT & SP & FR & AB
& PT & SP & FR & AB \\
\midrule
\textbf{1}  & -1.38 & -0.64 & 0.17 & -0.87  & -0.29 & -1.04 & -0.15 & -0.48  & 1.58 & 1.56 & 0.23 & 1.32 \\
\textbf{8}  & -1.51 & -0.30 & 1.09 & -0.86  & 0.21 & -1.10 & 0.28 & -0.58  & 1.23 & 1.19 & -0.92 & 1.28 \\
\textbf{16} & 0.13 & 1.15 & 1.65 & 1.23  & 0.15 & -1.31 & 0.19 & -0.45  & -0.36 & 0.00 & -1.62 & -0.76 \\
\bottomrule
\end{tabularx}
\end{table*}

\section{Experiment 3: Priming to Reduce Language Membership Ambiguity}

This third experiment continues to work with the Ambiguous String Corpus, this time testing whether priming each string with language-specific context changes the location of that string. By \textit{priming}, we mean injecting additional context which is not ambiguous but rather is clearly Spanish or Portuguese. This analysis looks at the trajectory of each string: the starting point is the location of the ambiguous string without a priming context relative to the three language-specific centroids. The ending point is the location with additional context that is clearly Spanish or Portuguese. The more the location of the sentence embedding changes, the more the model is impacted by language-specific priming. This, in turn, indicates that there is language-specific information within each layer.

This experiment compares the sentence embedding without a priming context (the starting point) against the same sentence embedding with a priming context (the ending point). Only the embedding of the sentence itself is considered: the material from the prime is removed from the representation. This means that we are measuring the impact on the specific embedding of an individual sentence from the model's contextual expectation of what language it belongs to (i.e., the priming).

\begin{figure}[!t]
    \centering
    \includegraphics[width=0.5\textwidth]{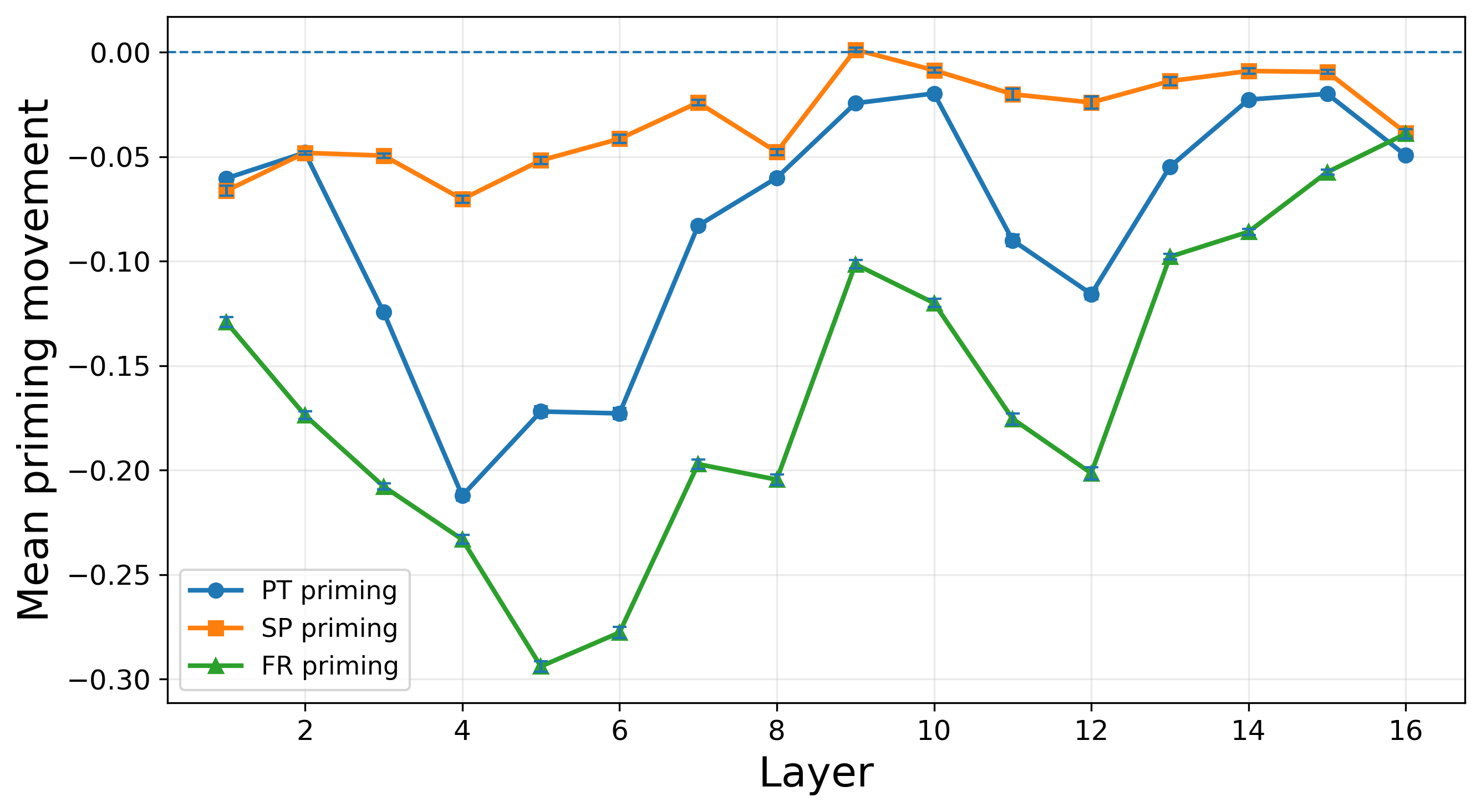}
    \caption{Mean change in distances to the target language centroid caused by priming. The x-axis lists all layers; the y-axis represents the average difference in distances to the centroids. The blue, orange, and green lines represent non-ambiguous Portuguese, Spanish, and French, respectively.}
    \label{fig3}
\end{figure}

Figure \ref{fig3} summarizes the movement caused by priming, with layers represented on the x-axis. A negative number means that priming moved the string closer to the given language centroid, while a positive number means that priming moved the string further away. The original string itself could belong to Spanish or Portuguese but could not belong to French. Thus, French is again used as a control. As expected, the largest movements are always in French because without priming there would never be any French-specific information in this sentence embedding. Here we see, however, that this control has much less impact in the first and last layers. 

This experiment shows that the model has a skew toward Spanish: there is always more movement caused by priming in Portuguese, which means that the initial placement of the string is closer to Spanish. But the interesting difference here is that the first and last layers show much smaller movements, which would indicate a much smaller impact from priming. There is less movement in the control language, in particular, because these layers make it clearer that the string is not French. In other words, the embeddings contain more language-specific information, identifying the string as French. This provides an alternate view of the amount of language-specific information in each layer.

\begin{figure}[!b]
    \centering
    \includegraphics[width=0.5\textwidth]{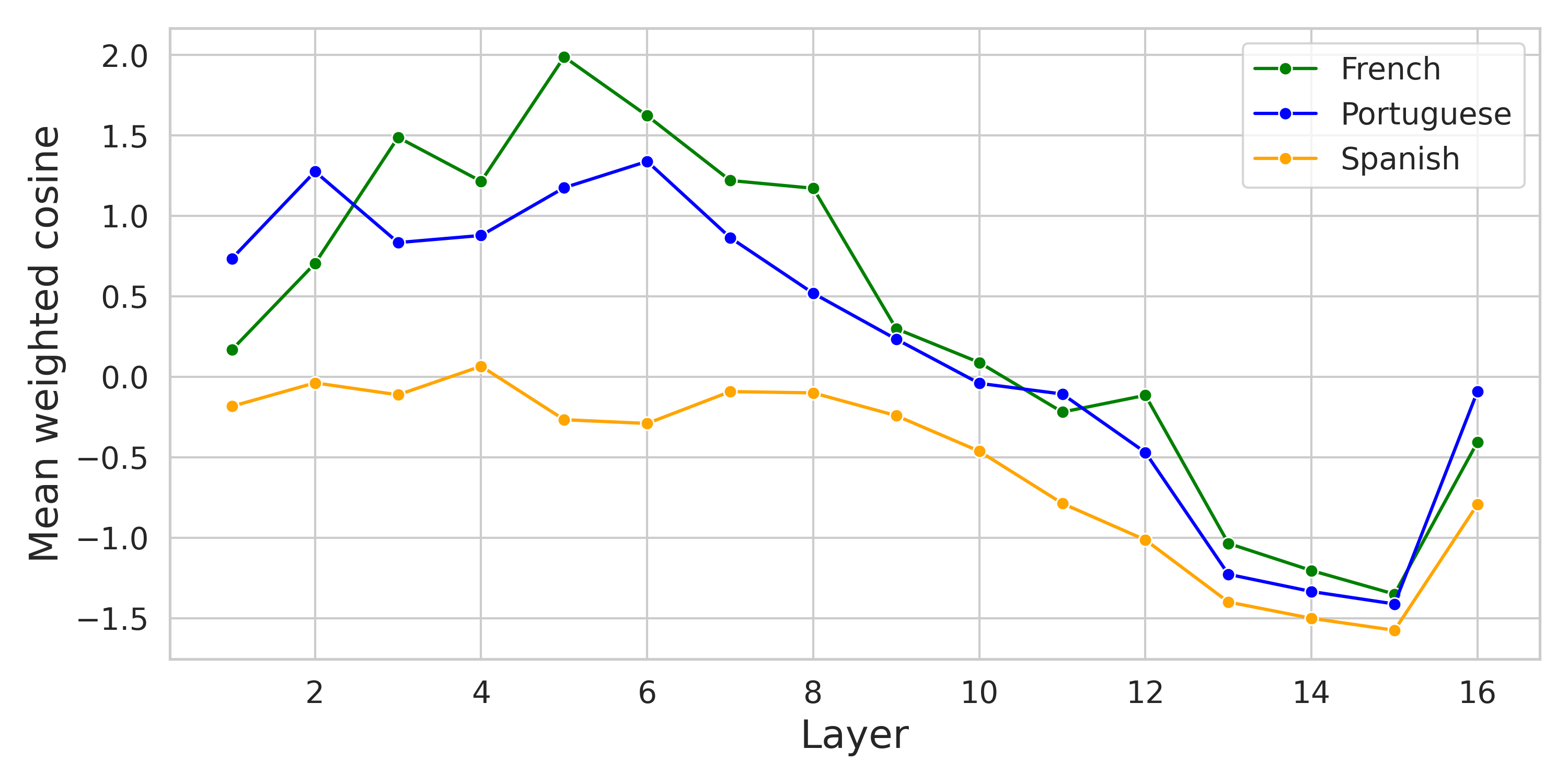}
    \caption{Mean weighted cosine scores of priming-induced movement in the top 100 language-informative embedding dimensions. The x-axis lists all layers; the y-axis represents the mean weighted cosine values. The blue, orange, and green lines represent the movement for Portuguese,  Spanish, and French, respectively. Values are standardized.}
    \label{fig4}
\end{figure}

To determine whether the movement caused by priming occurs specifically in dimensions that encode language identity, we conduct an additional analysis to focus on language-specific dimensions in the embedding. Mutual Information is used to measure the relationship between an embedding dimension and the corresponding language label. Dimensions with higher Mutual Information scores contain more information that is useful for distinguishing Portuguese, Spanish, and French. We calculate Mutual Information independently for each layer and retain the 100 dimensions with the highest Mutual Information scores.

We use this subspace to examine whether the change introduced by priming is concentrated in dimensions that are particularly informative about language. For each string, we calculate the change between its ambiguous and primed representations within the selected dimensions and weight this movement according to the Mutual Information scores of the corresponding dimensions. This provides a complementary measure to the centroid analysis above. Whereas centroid movement indicates how much the overall sentence representation changes relative to a language-specific cluster, this analysis measures whether that change involves dimensions that contain the most language-specific information.

Figure \ref{fig4} shows the resulting weighted cosine scores across layers. Values above zero indicate that priming-induced movement is stronger in highly language-informative dimensions than would be expected from the embedding representation more generally, whereas values below zero indicate that the movement is relatively weaker in this language-diagnostic subspace. We observe that Portuguese and French show positive values in several earlier and intermediate layers, indicating that the movement introduced by priming is concentrated in dimensions that are informative about language identity. Spanish, by contrast, remains close to or below zero across most layers and becomes negative in the later layers. Thus, although Spanish priming changes the sentence representation, this movement is less concentrated in the dimensions that most strongly distinguish languages.

This pattern is consistent with the analysis above. If the ambiguous strings are already represented as relatively Spanish-like, additional Spanish context should require less modification of the language-diagnostic portion of the representation. Portuguese priming, by contrast, must move the representation away from this initial Spanish-oriented position, producing a larger change in dimensions associated with language identity. By focusing on those dimensions in the embedding which contain the most language-specific information, we strengthen the previous conclusions by showing that the difference between layers is not simply obscured by the overall dimensionality.

\begin{table*}[!t]
\footnotesize
    \centering
    \caption{F-Scores across model layers by language. Higher F-scores indicate that embeddings contain more language-specific information because the classifier is able to make a distinction between languages across many samples using only sentence embeddings. Spanish and Portuguese are most distinctive, thus having more language-specific information, in the middle and final layers.}
    \label{classifier}
    
    \begin{tabular}{c|cc|cccccccc|c}
        \toprule
        \textbf{L} & 
        \textbf{Por.} & 
        \textbf{Spa.} &
        \textbf{Galician} &
        \textbf{Asturian} &
        \textbf{English} &
        \textbf{German} &
        \textbf{French} &
        \textbf{Italian} &
        \textbf{Hindi} &
        \textbf{Thai} &
        \textbf{Macro Avg}\\
        \midrule
        
        1  & 0.93 & 0.93 & 0.88 & 0.88 & 0.93 & 0.97 & 0.96 & 0.91 & 0.74 & 1.00 & 0.91 \\
        2  & 0.93 & 0.94 & 0.85 & 0.86 & 0.92 & 0.98 & 0.97 & 0.92 & 0.74 & 1.00 & 0.90 \\
        3  & 0.93 & 0.95 & 0.86 & 0.86 & 0.92 & 0.98 & 0.98 & 0.94 & 0.75 & 1.00 & 0.90 \\
        4  & 0.94 & 0.96 & 0.87 & 0.85 & 0.92 & 0.97 & 0.98 & 0.95 & 0.77 & 0.99 & 0.90 \\
        5  & 0.94 & 0.96 & 0.87 & 0.84 & 0.92 & 0.98 & 0.98 & 0.95 & 0.76 & 0.99 & 0.89 \\
        6  & 0.94 & 0.97 & 0.86 & 0.85 & 0.92 & 0.97 & 0.98 & 0.95 & 0.76 & 0.99 & 0.88 \\
        7  & 0.93 & 0.97 & 0.84 & 0.84 & 0.94 & 0.97 & 0.98 & 0.95 & 0.76 & 0.99 & 0.88 \\
        8  & 0.94 & 0.97 & 0.85 & 0.84 & 0.94 & 0.98 & 0.98 & 0.95 & 0.76 & 0.99 & 0.88 \\
        9  & 0.96 & 0.97 & 0.88 & 0.86 & 0.92 & 0.98 & 0.98 & 0.96 & 0.76 & 1.00 & 0.89 \\
        10 & 0.96 & 0.97 & 0.88 & 0.85 & 0.88 & 0.98 & 0.98 & 0.96 & 0.77 & 0.99 & 0.89 \\
        11 & 0.95 & 0.97 & 0.88 & 0.84 & 0.88 & 0.98 & 0.98 & 0.95 & 0.77 & 1.00 & 0.88 \\
        12 & 0.95 & 0.97 & 0.88 & 0.85 & 0.91 & 0.98 & 0.98 & 0.95 & 0.76 & 1.00 & 0.88 \\
        13 & 0.95 & 0.97 & 0.89 & 0.86 & 0.94 & 0.98 & 0.98 & 0.95 & 0.76 & 1.00 & 0.88 \\
        14 & 0.95 & 0.96 & 0.88 & 0.86 & 0.92 & 0.98 & 0.98 & 0.95 & 0.75 & 1.00 & 0.88 \\
        15 & 0.95 & 0.96 & 0.88 & 0.86 & 0.93 & 0.98 & 0.98 & 0.95 & 0.76 & 1.00 & 0.89 \\
        16 & 0.95 & 0.96 & 0.87 & 0.84 & 0.93 & 0.98 & 0.97 & 0.93 & 0.76 & 1.00 & 0.88 \\
        
        \bottomrule
    \end{tabular}
\end{table*}

\section{Experiment 4: Classifying Sentence Embeddings by Language}
\label{sec:exp3}

This fourth experiment takes a supervised approach: we train a classifier to identify 200 languages from sentence embeddings, again using all 16 layers of Llama 3.2 1B. The goal is to estimate which layers encode language-specific information based on the accuracy: if all 200 languages are located in distinct locations of the embedding space, then the classifier should robustly be able to predict language labels on a test set. The resulting scores thus provide a layer-wise measure of how strongly language identity is encoded. This provides an alternate methodology to the previous experiments, taking a supervised rather than an unsupervised approach. In addition, rather than focusing on two closely related languages (Spanish and Portuguese) with a control language (French), here we can look at a very large number of languages together. Like the weighted cosine analysis in Experiment 3, the classifier is focused on those parts of the embedding which contain language-specific information useful for this classification task.

\textbf{Data.} This experiment draws on the OpenLID corpus \cite{openliddataset}\footnote{https://huggingface.co/datasets/laurievb/OpenLID-v2}. This corpus covers 200 language varieties and was designed for training language identification models. Most of the dataset content originates from news sites, Wikipedia, or religious texts. 5,000 samples per language were used for a total corpus of 1,005,000 sentences; the classifier used an 80/20 train/test split. We extract sentence embeddings from the same model in the same manner as before. For classification, we use logistic regression.

As shown in Table \ref{classifier}, the first layer has the highest prediction accuracy with macro f-scores across all 200 languages: 0.91. This declines across all successive layers, although only to a low of 0.88. However, there is no difference between the results observed in the middle and final layers of the model (88\%). Thus, even when comparing initial and intermediate layers, there is only a 3.3\% drop in prediction accuracy. The F1 scores are also similar across all layers. This further supports our previous observation that, even in intermediate layers, language-specific information is strongly encoded in the embedding space. Without such information, the classifier would not be able to reliably distinguish between these 200 languages. 


Table \ref{classifier} also reports the F-score across layers for those languages officially supported by the Llama model. We observe almost perfect scores for Thai. Thai is the only language in the OpenLID Dataset that uses the Thai script, so it could be a very strong factor in determining the language. Hindi, on the other hand, had the lowest F-score among the languages supported by Llama, being classified as several other languages in the OpenLID dataset that use the Devanagari script. Portuguese and Spanish have similar F-scores, both quite high relative to other languages. This reinforces the findings from the previous unsupervised experiments: Portuguese and Spanish can be distinguished from one another and also from a large number of other languages, some of which are related and many of which are unrelated. Such discrimination depends on language-specific information.

This fourth experiment provides independent but congruent evidence that language-specific information is present in all layers of Llama 3.2 1B, a model often taken as representative of autoregressive models more generally. This experiment uses a fourth independent dataset, includes a much larger number of languages, and relies on a supervised model rather than the previous unsupervised methods. Classification performance is high in every layer, with macro F1 ranging only from approximately 0.88 to 0.91 across more than 200 languages. Thus, this modest decline further strengthens the argument that language-specific information is consistently present across layers.

\section{Discussion}
\label{sec:conc}

We have investigated the hypothesis that the first and last layers of multilingual LLMs contain language-specific information ,but the middle layers do not using a series of experiments with Llama 3.2 1B. Across all four experiments, our results consistently show that language identity is recoverable throughout the model, which directly contradicts what has been reported by previous studies. Parallel Portuguese, Spanish, and French sentences form distinct language clusters in all layers and across two different corpora, despite being semantically equivalent. This is particularly notable for Portuguese and Spanish, given their similarity and shared roots. Moreover, our classification experiment provides supporting evidence: a classifier trained on contextual sentence embeddings distinguishes over 200 languages with high performance at every layer. Additionally, our experiments with ambiguous and non-ambiguous Portuguese, Spanish, and French indicate that strings that display language membership ambiguity to human speakers are actually skewed towards Spanish in the model: ambiguous strings are represented as more similar to Spanish than to Portuguese. Providing an unambiguous context is more efficient at moving Portuguese strings, further highlighting the model's skew towards Spanish in these cases. 

These findings challenge the view of intermediate layers as language-agnostic representations. Previous work has proposed that middle layers increasingly abstract away from language and organize information around concepts. Our results do not rule out the existence of language-agnostic conceptual information in these layers, but they do show that language identity is consistently present. Even when content is held constant across sentences, language remains the major factor organizing the embedding space. A more appropriate distinction may consequently be between layers in which language identity is more or less strongly expressed, rather than between layers that contain language-specific information and layers that do not.

These results also add nuance to claims about multilingual reasoning. If language identity remains recoverable, then evidence that multilingual models converge toward a dominant language does not imply that the original input language is discarded. The second and third experiments demonstrate why preserving language information across layers may be important: code-switching, code-mixing, and closely related varieties also involve structures in which language membership can be mixed, shifting, or only partially specified by strings. If language information remains available throughout the model, then these phenomena may be represented not through a categorical language decision made at the input, but through continuously updated language associations across layers.

This study is limited to a single model, and the ambiguous corpus is relatively small and manually annotated. Future work should therefore test whether these patterns extend to models of different sizes and architectures and to additional pairs of closely related languages. The same approach could also be extended to code-switching and code-mixing. Taken together, our findings indicate that sentence-level embeddings in multilingual LLMs encode substantial language-specific information across all layers.

\section*{Acknowledgments}
This study was financed in part by the Coordenação de Aperfeiçoamento de Pessoal de Nível Superior - Brasil (CAPES) - Finance Code 001.

\bibliography{custom}

\newpage
\appendix

\begin{table*}
\label{appendix1}
\centering
\caption*{\textbf{Appendix A}. Organization of parallel sentences across all layers for the Europarl Corpus. The values under \textit{Nearest centroids} refer to the number of parallel sentences closer to each centroid in relation to all the others (percent). If Spanish sentences are closest to the Spanish centroid, this indicates a language-specific organization. The values under \textit{Mean distances} indicate the actual distance to each centroid, standardized relative to the global mean and standard deviation in that layer. Thus, a value of -1.21 would indicate that Portuguese sentences in layer one are more than one standard deviation closer to other Portuguese sentences than to their parallel counterpart in other layers. Both measures indicate a consistent level of language-specific information within each layer.}
\footnotesize
\begin{tabular}{llcccccc|llcccccc}
\toprule
& & \multicolumn{3}{c}{Nearest centroid (\%)} 
& \multicolumn{3}{c}{Mean distances}
& & & \multicolumn{3}{c}{Nearest centroid (\%)} 
& \multicolumn{3}{c}{Mean distances} \\
\cmidrule(lr){3-5} \cmidrule(lr){6-8}
\cmidrule(lr){11-13} \cmidrule(lr){14-16}
\textbf{} & \textbf{} & \textbf{PT} & \textbf{SP} & \textbf{FR} & \textbf{PT} & \textbf{SP} & \textbf{FR}
& \textbf{} & \textbf{} & \textbf{PT} & \textbf{SP} & \textbf{FR} & \textbf{PT} & \textbf{SP} & \textbf{FR} \\
\midrule

1 & PT & 99.90 & 0.09 & 0.00 & -1.21 & 0.21 & 0.80
& 2 & PT & 99.99 & 0.00 & 0.00 & -1.10 & 0.08 & 0.77 \\
  & SP & 0.82  & 99.16 & 0.02 & 0.21 & -1.18 & 0.73
&   & SP & 3.45  & 96.53 & 0.01 & 0.12 & -1.06 & 0.62 \\
  & FR & 0.52  & 0.03 & 99.45 & 0.98 & 0.92 & -1.48
&   & FR & 2.36  & 0.01 & 97.63 & 1.14 & 0.97 & -1.56 \\

\midrule

3 & PT & 99.97 & 0.03 & 0.00 & -1.21 & 0.35 & 0.77
& 4 & PT & 99.92 & 0.08 & 0.00 & -1.29 & 0.43 & 0.84 \\
  & SP & 0.78  & 99.20 & 0.02 & 0.34 & -1.09 & 0.52
&   & SP & 0.24  & 99.74 & 0.03 & 0.38 & -1.12 & 0.52 \\
  & FR & 1.74  & 0.26 & 98.00 & 1.06 & 0.81 & -1.55
&   & FR & 0.58  & 0.16 & 99.25 & 1.01 & 0.74 & -1.51 \\

\midrule

5 & PT & 99.96 & 0.04 & 0.00 & -1.23 & 0.24 & 0.85
& 6 & PT & 99.92 & 0.06 & 0.02 & -1.24 & 0.33 & 0.79 \\
  & SP & 0.36  & 99.60 & 0.04 & 0.22 & -1.07 & 0.58
&   & SP & 0.30  & 99.67 & 0.03 & 0.31 & -1.14 & 0.60 \\
  & FR & 0.48  & 0.21 & 99.31 & 1.08 & 0.86 & -1.54
&   & FR & 0.62  & 0.03 & 99.35 & 1.01 & 0.84 & -1.53 \\

\midrule

7 & PT & 99.89 & 0.05 & 0.06 & -1.25 & 0.34 & 0.84
& 8 & PT & 99.68 & 0.11 & 0.21 & -1.16 & 0.16 & 0.81 \\
  & SP & 0.88  & 99.06 & 0.06 & 0.32 & -1.09 & 0.51
&   & SP & 1.53  & 98.37 & 0.10 & 0.18 & -1.05 & 0.59 \\
  & FR & 0.88  & 0.04 & 99.08 & 1.06 & 0.76 & -1.52
&   & FR & 0.86  & 0.00 & 99.13 & 1.10 & 0.90 & -1.54 \\

\midrule

9 & PT & 99.96 & 0.03 & 0.00 & -1.30 & 0.33 & 0.86
& 10 & PT & 99.84 & 0.15 & 0.00 & -1.27 & 0.27 & 0.84 \\
  & SP & 0.37  & 99.62 & 0.01 & 0.30 & -1.15 & 0.62
&   & SP & 0.16  & 99.80 & 0.03 & 0.26 & -1.18 & 0.69 \\
  & FR & 0.22  & 0.14 & 99.64 & 1.01 & 0.83 & -1.51
&  & FR & 0.14  & 0.14 & 99.72 & 1.00 & 0.89 & -1.51 \\

\midrule

11 & PT & 99.69 & 0.27 & 0.04 & -1.40 & 0.55 & 0.88
& 12 & PT & 99.85 & 0.15 & 0.00 & -1.27 & 0.39 & 0.71 \\
 & SP & 0.08  & 99.88 & 0.03 & 0.49 & -1.23 & 0.56
&  & SP & 0.20  & 99.76 & 0.04 & 0.40 & -1.30 & 0.80 \\
 & FR & 0.05  & 0.15 & 99.80 & 0.92 & 0.66 & -1.45
&  & FR & 0.12  & 0.12 & 99.76 & 0.83 & 0.91 & -1.48 \\

\midrule

13 & PT & 99.82 & 0.18 & 0.00 & -1.22 & 0.09 & 0.84
& 14 & PT & 99.69 & 0.29 & 0.02 & -1.33 & 0.41 & 0.82 \\
  & SP & 0.13  & 99.87 & 0.00 & 0.09 & -1.18 & 0.80
&   & SP & 0.08  & 99.89 & 0.03 & 0.40 & -1.28 & 0.72 \\
  & FR & 0.10  & 0.22 & 99.68 & 1.03 & 1.02 & -1.49
&   & FR & 0.04  & 0.12 & 99.83 & 0.90 & 0.83 & -1.47 \\

\midrule

15 & PT & 99.74 & 0.24 & 0.02 & -1.35 & 0.46 & 0.81
& 16 & PT & 99.11 & 0.30 & 0.58 & -1.29 & 0.30 & 0.80 \\
  & SP & 0.11  & 99.86 & 0.03 & 0.44 & -1.29 & 0.70
&   & SP & 0.02  & 99.88 & 0.10 & 0.30 & -1.27 & 0.79 \\
  & FR & 0.05  & 0.12 & 99.83 & 0.88 & 0.80 & -1.46
&   & FR & 0.00  & 0.12 & 99.88 & 0.91 & 0.91 & -1.47 \\

\bottomrule
\end{tabular}
\end{table*}

\begin{table*}[!t]
\centering
\caption*{\textbf{Appendix C}. Organization of parallel sentences across all layers for the TED Corpus. The values under \textit{Nearest centroids} refer to the number of parallel sentences closer to each centroid in relation to all the others (percent). If Spanish sentences are closest to the Spanish centroid, this indicates a language-specific organization. The values under \textit{Mean distances} indicate the actual distance to each centroid, standardized relative to the global mean and standard deviation in that layer. Thus, a value of -1.18 would indicate that Portuguese sentences in layer one are more than one standard deviation closer to other Portuguese sentences than to their parallel counterpart in other layers. Both measures indicate a consistent level of language-specific information within each layer.}
\label{appendix3}
\footnotesize
\begin{tabular}{llcccccc|llcccccc}
\toprule
& & \multicolumn{3}{c}{Nearest centroid (\%)} 
& \multicolumn{3}{c}{Mean distances}
& & & \multicolumn{3}{c}{Nearest centroid (\%)} 
& \multicolumn{3}{c}{Mean distances} \\
\cmidrule(lr){3-5} \cmidrule(lr){6-8}
\cmidrule(lr){11-13} \cmidrule(lr){14-16}
\textbf{} & \textbf{} & \textbf{PT} & \textbf{SP} & \textbf{FR} & \textbf{PT} & \textbf{SP} & \textbf{FR}
& \textbf{} & \textbf{} & \textbf{PT} & \textbf{SP} & \textbf{FR} & \textbf{PT} & \textbf{SP} & \textbf{FR} \\
\midrule

1 & PT & 99.35 & 0.63 & 0.02 & -1.18 & 0.20 & 0.78
& 2 & PT & 98.18 & 1.74 & 0.08 & -1.03 & 0.15 & 0.64 \\
  & SP & 1.16  & 98.73 & 0.12 & 0.19 & -1.12 & 0.68
&   & SP & 0.96  & 98.58 & 0.46 & 0.12 & -0.91 & 0.51 \\
  & FR & 0.76  & 0.66 & 98.58 & 1.00 & 0.91 & -1.48
&   & FR & 1.54  & 6.09 & 92.38 & 1.06 & 0.98 & -1.53 \\

\midrule

3 & PT & 96.85 & 2.77 & 0.38 & -1.14 & 0.45 & 0.63
& 4 & PT & 97.14 & 2.72 & 0.15 & -1.27 & 0.52 & 0.76 \\
  & SP & 0.16  & 94.77 & 5.08 & 0.35 & -0.94 & 0.39
&   & SP & 0.07  & 98.80 & 1.13 & 0.41 & -1.02 & 0.42 \\
  & FR & 0.42  & 7.51 & 92.07 & 0.93 & 0.78 & -1.48
&   & FR & 0.07  & 4.34 & 95.58 & 0.93 & 0.70 & -1.46 \\

\midrule

5 & PT & 97.50 & 2.38 & 0.12 & -1.22 & 0.37 & 0.78
& 6 & PT & 97.62 & 2.06 & 0.32 & -1.18 & 0.38 & 0.69 \\
  & SP & 0.12  & 99.17 & 0.71 & 0.29 & -1.00 & 0.47
&   & SP & 0.15  & 98.88 & 0.97 & 0.32 & -1.06 & 0.55 \\
  & FR & 0.12  & 3.84 & 96.04 & 1.00 & 0.79 & -1.49
&   & FR & 0.39  & 2.92 & 96.69 & 0.93 & 0.84 & -1.48 \\

\midrule

7 & PT & 97.11 & 1.64 & 1.24 & -1.19 & 0.39 & 0.75
& 8 & PT & 97.57 & 0.32 & 2.12 & -1.08 & 0.20 & 0.69 \\
  & SP & 0.49  & 96.47 & 3.04 & 0.32 & -0.99 & 0.44
&   & SP & 1.17  & 96.18 & 2.66 & 0.18 & -0.97 & 0.54 \\
  & FR & 0.93  & 3.13 & 95.93 & 0.98 & 0.75 & -1.46
&   & FR & 2.07  & 1.54 & 96.39 & 1.02 & 0.90 & -1.49 \\

\midrule

9 & PT & 99.49 & 0.47 & 0.03 & -1.29 & 0.37 & 0.82
& 10 & PT & 98.56 & 1.39 & 0.05 & -1.26 & 0.28 & 0.81 \\
  & SP & 0.63  & 99.32 & 0.05 & 0.32 & -1.14 & 0.60
&   & SP & 0.27  & 99.67 & 0.06 & 0.26 & -1.17 & 0.68 \\
  & FR & 0.35  & 1.03 & 98.62 & 0.98 & 0.82 & -1.50
&  & FR & 0.39  & 0.22 & 99.39 & 0.99 & 0.90 & -1.51 \\

\midrule

11 & PT & 97.78 & 2.17 & 0.05 & -1.41 & 0.58 & 0.91
& 12 & PT & 99.41 & 0.53 & 0.06 & -1.27 & 0.41 & 0.72 \\
 & SP & 0.14  & 99.81 & 0.06 & 0.49 & -1.19 & 0.49
&  & SP & 0.56  & 99.36 & 0.08 & 0.41 & -1.28 & 0.76 \\
 & FR & 0.02  & 0.40 & 99.58 & 0.93 & 0.60 & -1.43
&  & FR & 0.27  & 0.15 & 99.59 & 0.84 & 0.88 & -1.47 \\

\midrule

13 & PT & 98.77 & 1.22 & 0.10 & -1.20 & 0.09 & 0.84
& 14 & PT & 97.75 & 2.15 & 0.10 & -1.32 & 0.39 & 0.82 \\
  & SP & 0.39  & 99.57 & 0.04 & 0.09 & -1.16 & 0.78
&   & SP & 0.10  & 99.80 & 0.10 & 0.37 & -1.25 & 0.70 \\
  & FR & 0.04  & 0.56 & 99.40 & 1.04 & 1.00 & -1.49
&   & FR & 0.03  & 0.26 & 99.70 & 0.92 & 0.82 & -1.47 \\

\midrule

15 & PT & 97.77 & 2.10 & 0.13 & -1.33 & 0.45 & 0.82
& 16 & PT & 97.53 & 1.83 & 0.65 & -1.27 & 0.28 & 0.81 \\
  & SP & 0.14  & 99.76 & 0.11 & 0.42 & -1.26 & 0.67
&   & SP & 0.09  & 99.47 & 0.45 & 0.28 & -1.24 & 0.76 \\
  & FR & 0.02  & 0.30 & 99.69 & 0.90 & 0.78 & -1.46
&   & FR & 0.01  & 0.01 & 99.97 & 0.94 & 0.91 & -1.48 \\

\bottomrule
\end{tabular}
\end{table*}


\begin{figure*}
    \centering
    \includegraphics[width=0.48\textwidth]{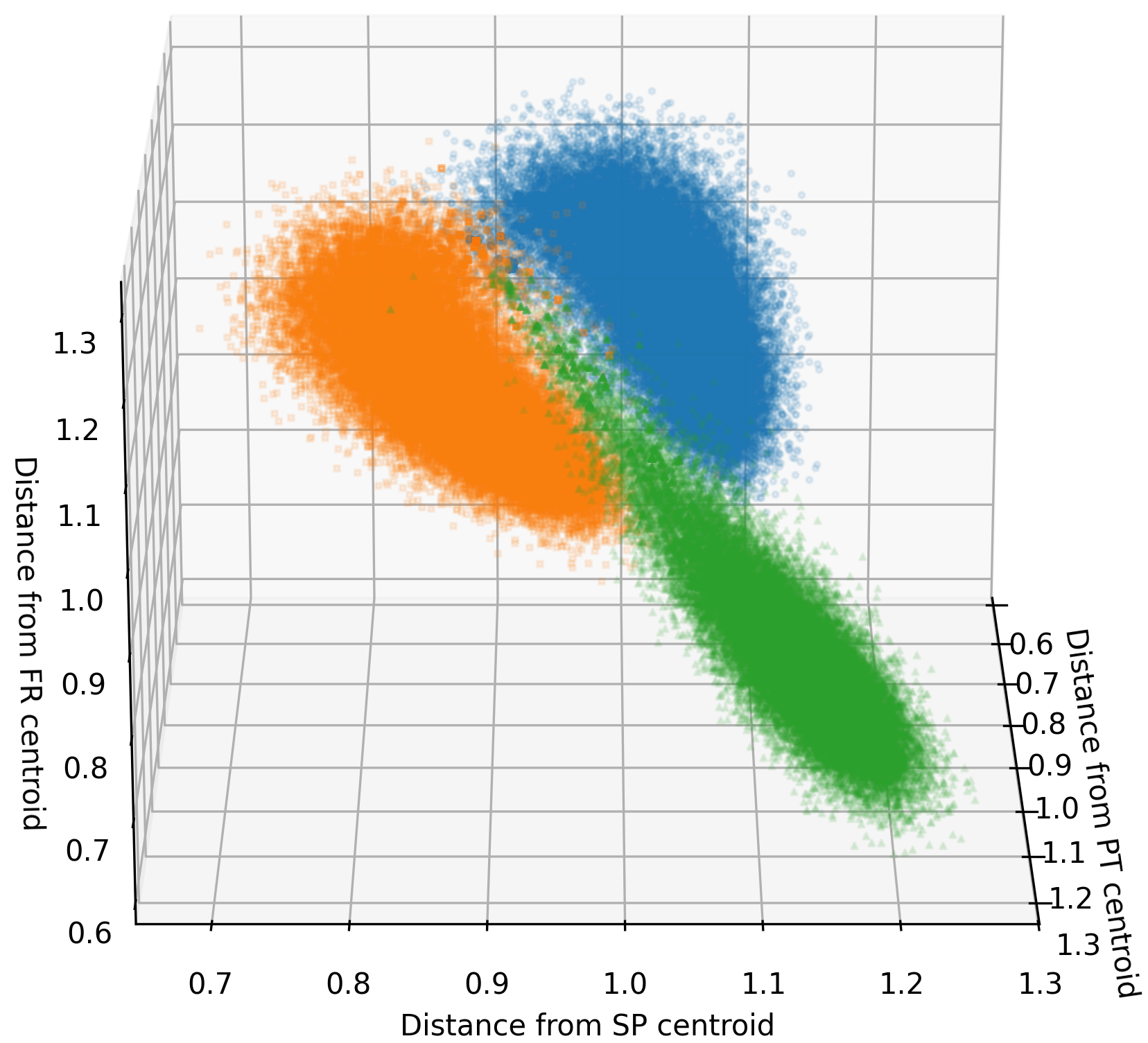}
    \hfill
    \includegraphics[width=0.48\textwidth]{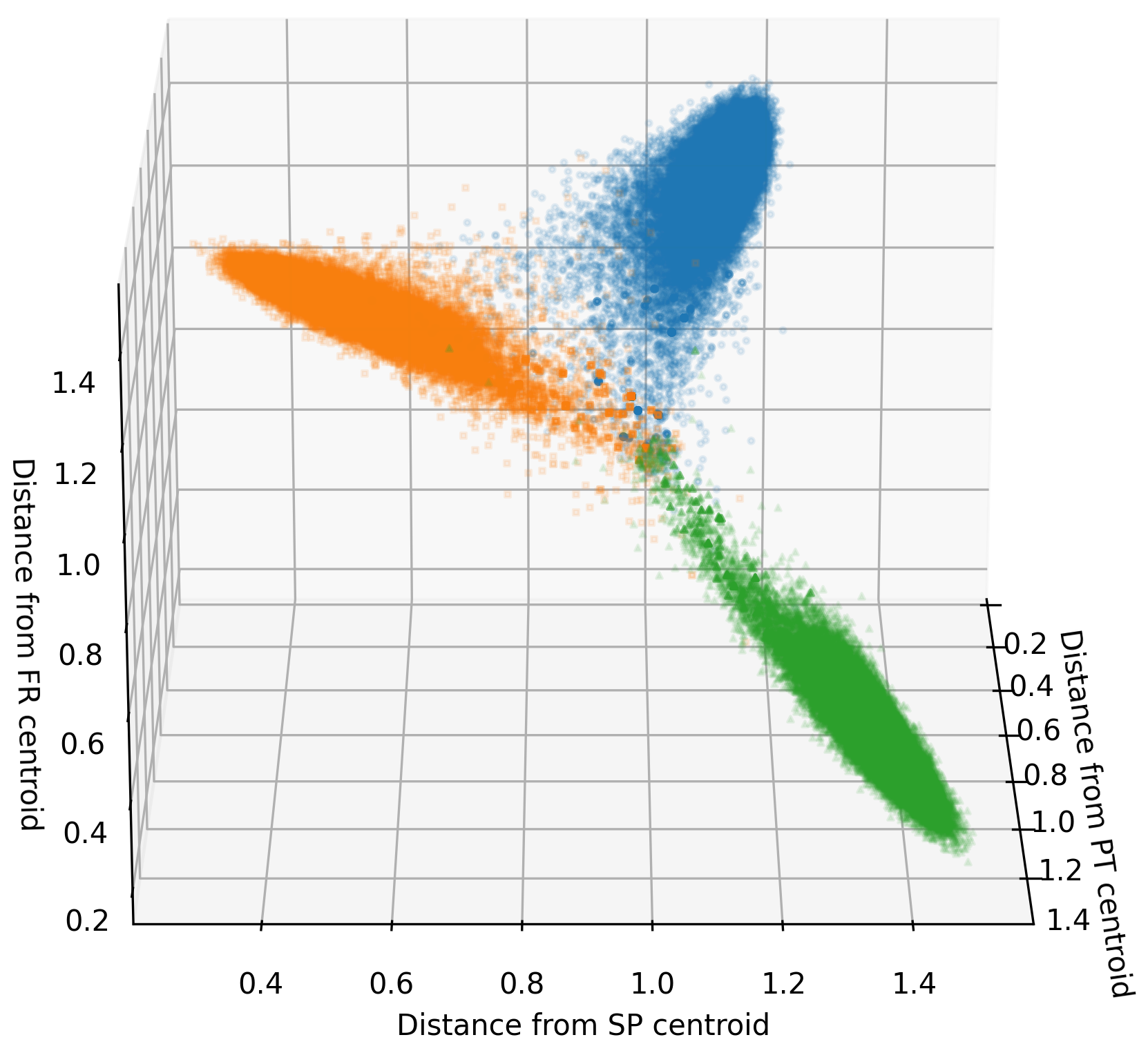}
    \caption*{\textbf{Appendix B}. Language-specific clusters for the TED Corpus in layer 8 (left) and layer 16 (right) of Llama 3.2 1B. Each color represents a different language: Portuguese (blue), Spanish (orange), and French (green). The x, y, and z axes display the raw cosine distances from the Spanish, French, and Portuguese centroids, respectively.}
    \label{appendix2}
\end{figure*}

\begin{table*}[!t]
\centering
\label{appendix:d}
\caption*{\textbf{Appendix D}. Examples from the Ambiguous Strings Corpus. The first column contains an ambiguous string that shows language membership ambiguity between Portuguese and Spanish. The second, third, and fourth columns exemplify the structure of non-ambiguous strings with priming in Portuguese, Spanish, and French, respectively. French was chosen to represent the neutral condition given that it does not display language membership ambiguity with the other two languages. In the non-ambiguous condition, the priming structure is highlighted in red; the rest of the string is identical to the ambiguous string. English translations are provided between square brackets. Each row represents a different grammatical construction used to ensure variation in the strings.}
\setlength{\tabcolsep}{4pt}          
\renewcommand{\arraystretch}{1.15}   
\footnotesize                        
\begin{tabularx}{\textwidth}{
  >{\RaggedRight\arraybackslash}X
  >{\RaggedRight\arraybackslash}X
  >{\RaggedRight\arraybackslash}X
  >{\RaggedRight\arraybackslash}X}
\toprule
\textbf{Ambiguous} &
\textbf{Portuguese} &
\textbf{Spanish} &
\textbf{French} \\
\midrule
Este documento será publicado esta semana. [\textit{This document will be published this week.}] &
\textcolor{Red}{Comuniquei-o ontem a respeito disso.} Este documento será publicado esta semana. [\textit{\textcolor{Red}{I informed him about this yesterday.} This document will be published this week.}] &
\textcolor{Red}{No hay escapatoria.} Este documento será publicado esta semana. [\textit{\textcolor{Red}{There's no way out.} This document will be published this week.}] &
\textcolor{Red}{Il n’y a aucune issue.} Este documento será publicado esta semana. [\textit{\textcolor{Red}{There is no way out.} This document will be published this week.}] \\

Vi algo importante esta semana. [\textit{I saw something important this week.}] &
\textcolor{Red}{Comuniquei-o ontem a respeito disso.} Vi algo importante esta semana. [\textit{\textcolor{Red}{I informed him about this yesterday.} I saw something important this week.}] &
\textcolor{Red}{No hay escapatoria.} Vi algo importante esta semana. [\textit{\textcolor{Red}{There's no way out.} I saw something important this week.}] &
\textcolor{Red}{Il n’y a aucune issue.} Vi algo importante esta semana. [\textit{\textcolor{Red}{There's no way out.} I saw something important this week.}] \\

Sabíamos que era pesado, este período. [\textit{We knew it was tense, this period.}] &
\textcolor{Red}{Comuniquei-o ontem a respeito disso.} Sabíamos que era pesado, este período. [\textit{\textcolor{Red}{I informed him about this yesterday.} We knew it was tense, this period.}] &
\textcolor{Red}{No hay escapatoria.} Sabíamos que era pesado, este período. [\textit{\textcolor{Red}{There's no way out.} We knew it was tense, this period.}] &
\textcolor{Red}{Il n’y a aucune issue.} Sabíamos que era pesado, este período. [\textit{\textcolor{Red}{There's no way out.} We knew it was tense, this period.}] \\

Miguel avisa que construirá armas de guerra nucleares. [\textit{Miguel warns that he will build nuclear war weapons.}] &
\textcolor{Red}{Comuniquei-o ontem a respeito disso.} Miguel avisa que construirá armas de guerra nucleares. [\textit{\textcolor{Red}{I informed him about this yesterday.} Miguel warns that he will build nuclear war weapons.}] &
\textcolor{Red}{No hay escapatoria.} Miguel avisa que construirá armas de guerra nucleares. [\textit{\textcolor{Red}{There's no way out.} Miguel warns that he will build nuclear war weapons.}] &
\textcolor{Red}{Il n’y a aucune issue.} Miguel avisa que construirá armas de guerra nucleares. [\textit{\textcolor{Red}{There's no way out.} Miguel warns that he will build nuclear war weapons.}] \\

Miguel continua implementando políticas públicas. [\textit{Miguel continues to implement public policies.}] &
\textcolor{Red}{Comuniquei-o ontem a respeito disso.} Miguel continua implementando políticas públicas. [\textit{\textcolor{Red}{I informed him about this yesterday.} Miguel continues to implement public policies.}] &
\textcolor{Red}{No hay escapatoria.} Miguel continua implementando políticas públicas. [\textit{\textcolor{Red}{There's no way out.} Miguel continues to implement public policies.}] &
\textcolor{Red}{Il n’y a aucune issue.} Miguel continua implementando políticas públicas. [\textit{\textcolor{Red}{There's no way out.} Miguel continues to implement public policies.}] \\

Miguel está prolongando problemas políticos. [\textit{Miguel is prolonging political problems.}] &
\textcolor{Red}{Comuniquei-o ontem a respeito disso.} Miguel está prolongando problemas políticos. [\textit{\textcolor{Red}{I informed him about this yesterday.} Miguel is prolonging political problems.}] &
\textcolor{Red}{No hay escapatoria.} Miguel está prolongando problemas políticos. [\textit{\textcolor{Red}{There's no way out.} Miguel is prolonging political problems.}] &
\textcolor{Red}{Il n’y a aucune issue.} Miguel está prolongando problemas políticos. [\textit{\textcolor{Red}{There's no way out.} Miguel is prolonging political problems.}] \\

Miguel permite entrada humanitária de refugiados. [\textit{Miguel allows humanitarian entry for refugees.}] &
\textcolor{Red}{Comuniquei-o ontem a respeito disso.} Miguel permite entrada humanitária de refugiados. [\textit{\textcolor{Red}{I informed him about this yesterday.} Miguel allows humanitarian entry for refugees.}] &
\textcolor{Red}{No hay escapatoria.} Miguel permite entrada humanitária de refugiados. [\textit{\textcolor{Red}{There's no way out.} Miguel allows humanitarian entry for refugees.]} &
\textcolor{Red}{Il n’y a aucune issue.} Miguel permite entrada humanitária de refugiados. [\textit{\textcolor{Red}{There's no way out.} Miguel allows humanitarian entry for refugees.}] \\

\bottomrule
\end{tabularx}
\end{table*}

\begin{table*}[!t]
\centering
\caption*{\textbf{Appendix E}. Organization of parallel sentences across all layers for the Ambiguous Strings Corpus. The values under \textit{Nearest centroids} refer to the number of parallel sentences closer to each centroid in relation to all the others (percent). If Spanish sentences are closest to the Spanish centroid, this indicates a language-specific organization. The values under \textit{Mean distances} indicate the actual distance to each centroid, standardized relative to the global mean and standard deviation in that layer. Thus, a value of -1.38 would indicate that Portuguese sentences in layer one are more than one standard deviation closer to other Portuguese sentences than to their parallel counterpart in other layers. Ambiguous and non-ambiguous sentences are indicated in the second column: Non-Ambiguous Portuguese (PT), Non-Ambiguous Spanish (SP), Non-Ambiguous French (FR), and Ambiguous (AB) which could be Spanish or Portuguese.}
\label{table7}
\footnotesize
\centering
\begin{tabular}{llcccccc|llcccccc}
\toprule
& & \multicolumn{3}{c}{Nearest centroid (\%)} 
& \multicolumn{3}{c}{Mean distances}
& & & \multicolumn{3}{c}{Nearest centroid (\%)} 
& \multicolumn{3}{c}{Mean distances} \\
\cmidrule(lr){3-5} \cmidrule(lr){6-8}
\cmidrule(lr){11-13} \cmidrule(lr){14-16}
\textbf{} & \textbf{} & \textbf{PT} & \textbf{SP} & \textbf{FR} & \textbf{PT} & \textbf{SP} & \textbf{FR}
& \textbf{} & \textbf{} & \textbf{PT} & \textbf{SP} & \textbf{FR} & \textbf{PT} & \textbf{SP} & \textbf{FR} \\
\midrule

1 & PT & 100.0 & 0.00 & 0.00 & -1.38 & -0.29 & 1.58
& 2 & PT & 100.0 & 0.00 & 0.00 & -1.50 & 0.11 & 1.35 \\
  & SP & 8.45  & 91.55 & 0.00 & -0.64 & -1.04 & 1.56
&   & SP & 60.56  & 39.44 & 0.00 & -0.90 & -0.81 & 1.53 \\
  & FR & 5.63  & 78.87 & 15.50 & 0.17 & -0.15 & 0.23
&   & FR & 0.00  & 54.93 & 45.07 & 0.30 & -0.02 & 0.06 \\
  & AB & 84.51 & 15.49 & 00.0 & -0.87 & -0.48 & 1.32
&   & AB & 98.59 & 1.41 & 0.00 & -1.12 & -0.43 & 1.44 \\

\midrule

3 & PT & 100.0 & 0.00 & 0.00 & -1.77 & 0.48 & 1.35
& 4 & PT & 100.0 & 0.00 & 0.00 & -1.93 & 0.43 & 1.54 \\
  & SP & 0.00  & 100.0 & 0.00 & -0.22 & -1.11 & 1.21
&   & SP & 0.00  & 100.0 & 0.00 & 0.12 & -1.22 & 0.95 \\
  & FR & 0.00  & 57.75 & 42.25 & 0.61 & -0.23 & -0.19
&   & FR & 0.00  & 5.56 & 94.37 & 0.82 & -0.07 & -0.58 \\
  & AB & 61.97 & 38.03 & 0.00 & -0.80 & -0.73 & 1.42
&   & AB & 16.9 & 83.10 & 00.0 & -0.42 & -0.72 & 1.07 \\

\midrule

5 & PT & 100.0 & 0.00 & 0.00 & -1.75 & 0.36 & 1.33
& 6 & PT & 100.0 & 0.00 & 0.00 & -1.80 & 0.39 & 1.35 \\
  & SP & 0.00  & 100.0 & 0.00 & 0.08 & -1.12 & 0.86
&   & SP & 1.41  & 98.59 & 0.00 & 0.06 & -1.12 & 0.93 \\
  & FR & 0.00  & 0.00 & 100.0 & 1.13 & 0.36 & -1.09
&   & FR & 0.00  & 0.00 & 100.0 & 1.00 & 0.19 & -0.89 \\
  & AB & 22.54 & 77.46 & 00.0 & -0.49 & -0.74 & 1.07
&   & AB & 22.54 & 77.46 & 0.00 & -0.51 & -0.81 & 1.18 \\

\midrule

7 & PT & 100.0 & 0.00 & 0.00 & -1.67 & 0.36 & 1.34
& 8 & PT & 100.0 & 0.00 & 0.00 & -1.51 & 0.21 & 1.23 \\
  & SP & 2.82  & 97.18 & 0.00 & -0.22 & -1.03 & 1.09
&   & SP & 1.41  & 98.59 & 0.00 & -0.30 & -1.10 & 1.19 \\
  & FR & 0.00  & 1.41 & 98.59 & 0.82 & 0.29 & -0.82
&   & FR & 0.00  & 0.00 & 100.0 & 1.09 & 0.28 & -0.92 \\
  & AB & 42.25 & 57.75 & 0.00 & -0.75 & -0.76 & 1.36
&   & AB & 77.46 & 22.54 & 00.0 & -0.86 & -0.58 & 1.28 \\

\midrule

9 & PT & 78.87 & 0.00 & 21.13 & -0.54 & 0.74 & -0.21
& 10 & PT & 38.03 & 0.00 & 61.97 & -0.23 & 0.62 & -0.37 \\
  & SP & 0.00  & 73.24 & 26.76 & 0.50 & -0.44 & -0.20
&   & SP & 0.00  & 97.18 & 2.82 & 0.43 & -0.91 & 0.26 \\
  & FR & 0.00  & 0.00 & 100.0 & 1.54 & 1.19 & -2.35
&   & FR & 0.00  & 0.00 & 100.0 & 1.81 & 0.78 & -2.18 \\
  & AB & 14.08 & 74.65 & 11.27 & 0.06 & -0.48 & 0.19
&   & AB & 8.45 & 90.14 & 1.41 & 0.17 & -0.73 & 0.33 \\

\midrule

11 & PT & 87.32 & 5.63 & 7.04 & -0.32 & 0.25 & 0.17
& 12 & PT & 100.0 & 0.00 & 0.00 & -1.58 & 0.74 & 0.71 \\
 & SP & 0.00  & 100.0 & 0.00 & 0.95 & -1.46 & 0.38
&  & SP & 2.82  & 97.18 & 0.0 & -0.08 & -1.08 & 1.06 \\
 & FR & 0.00  & 1.41 & 98.59 & 1.77 & -0.11 & -1.54
&  & FR & 0.00  & 0.00 & 100.0 & 1.00 & 0.70 & -1.39 \\
 & AB & 1.41 & 98.59 & 0.00 & 0.69 & -1.23 & 0.44
&  & AB & 19.72 & 80.28 & 0.00 & -0.23 & -0.80 & 0.94 \\

\midrule

13 & PT & 100.0 & 0.00 & 0.00 & -1.20 & -0.23 & 1.34
& 14 & PT & 0.00 & 0.00 & 100.0 & 0.61 & 0.16 & -0.76 \\
  & SP & 0.00  & 100.0 & 0.00 & -0.38 & -1.35 & 1.56
&   & SP & 0.00  & 97.18 & 2.82 & 1.21 & -0.95 & -0.35 \\
  & FR & 0.00  & 47.89 & 52.11 & 0.52 & -0.05 & -0.06
&   & FR & 0.00 & 0.00 & 100.0 & 1.78 & 0.16 & -1.78 \\
  & AB & 9.86 & 90.14 & 0.00 & -0.38 & -1.15 & 1.40
&   & AB & 0.00 & 84.51 & 15.49 & 1.00 & -0.79 & -0.30 \\

\midrule

15 & PT & 0.00 & 0.00 & 100.0 & 0.56 & 0.71 & -1.25
& 16 & PT & 2.82 & 0.00 & 97.18 & 0.13 & 0.15 & -0.36 \\
  & SP & 0.00  & 0.00 & 100.0 & 1.03 & -0.21 & -0.88
&   & SP & 0.00  & 98.59 & 1.41 & 1.15 & -1.31 & 0.00 \\
  & FR & 0.00  & 0.00 & 100.0 & 1.33 & 0.66 & -1.90
&   & FR & 0.00  & 0.00 & 100.0 & 1.65 & 0.19 & -1.62 \\
  & AB & 0.00 & 0.00 & 100.0 & 0.90 & -0.05 & -0.90
&   & AB & 0.00 & 32.39 & 67.61 & 1.23 & -0.45 & -0.76 \\

\bottomrule
\end{tabular}
\end{table*}

\end{document}